# A High-accuracy Calibration Method of Transient TSEPs for Power Semiconductor Devices

Qinghao Zhang, Wenrui Li and Pinjia Zhang*, *Senior Member, IEEE*

*Abstract*—The thermal sensitive electrical parameter (TSEP) method is crucial for enhancing the reliability of power devices through junction temperature monitoring. The TSEP method comprises three key processes: calibration, regression, and application. While significant efforts have been devoted to improving regression algorithms and increasing TSEP sensitivity to enhance junction temperature monitoring accuracy, these approaches have reached a bottleneck. In reality, the calibration method significantly influences monitoring accuracy, an aspect often overlooked in conventional TSEP methods. To address this issue, we propose a high-accuracy calibration method for transient TSEPs. *First*, a temperature compensation strategy based on thermal analysis is introduced to mitigate the temperature difference caused by load current during dual pulse tests. *Second*, the impact of stray parameters is analyzed to identify coupled parameters, which are typically neglected in existing methods. *Third*, it is observed that random errors follow a logarithm Gaussian distribution, covering a hidden variable. A neural network is used to obtain the junction temperature predictive model. The proposed calibration method is experimental validated in threshold voltage as an example. Compared with conventional calibration methods, the mean absolute error is reduced by over 30%. Moreover, this method does not require additional hardware cost and has good generalization.

## I. INTRODUCTION

Power electronic (PE) technology has been increasingly employed in industry, such as the new power generation system, electrified transportations, and the direct-current transmission. As the key and most vulnerable component of a PE system, the reliability of power semiconductor devices is of significance, since failures of them can result in considerable economic losses and safety incidents. After a prolonged research accumulation, the junction temperature ($T_j$) has been recognized as the key failure factor, as over 50% of device failures are caused by overheating or $T_j$ fluctuation [1]. To lower the thermal failure rate and enhance the reliability of PE devices, the $T_j$ monitoring technique was devised in the last century. After developments, the thermal sensitivity electrical parameter (TSEP) method has emerged as the most widely utilized $T_j$ monitoring technique, thanks to its advantages in high response speed, the straightforward modelling process, and the low difficulty in online implementation [2].

The TSEP method comprises three fundamental processes, namely calibration, regression and application [3]. Calibration involves measuring and sampling the data of $T_j$ and the corresponding TSEP feature $\boldsymbol{x} = [x_1, x_2,…]$, namely, the training data. Regression involves establishing a predictive model $\boldsymbol{F}$ of $T_j$ and $\boldsymbol{x}$, denoted as $T_j = \boldsymbol{F}(\boldsymbol{x})$. Subsequently, during normal operation of the target device, namely the application process, the junction temperature can be monitored as $T_{jm} = \boldsymbol{F}(\boldsymbol{x}^+)$, where $\boldsymbol{x}^+$ represents a new set of measured data.

Clearly, the performance of model $\boldsymbol{F}$ is the most crucial as it determines the accuracy of the $T_j$ monitoring method, which serves as the core evaluation index. Numerous efforts have been made to enhance the performance of model $\boldsymbol{F}$ and the monitoring accuracy including identifying a high-sensitivity TSEP and attempting improving the accuracy of regression algorithm for its establishment. However, both approaches have certain limitations.

The sensitivity of a TSEP is determined by its physical mechanisms, which can only be improved at a high cost. In the case of the classical on-state voltage drop ($V_{ON}$) or resistance ($R_{ON}$) method, the sensitivity is determined by the static operating point of the power device [4-5], controlled by the level of gate voltage ($V_g$). Lowering the high level of $V_g$ may enhance the sensitivity of $R_{ON}$ [6], but it also results in higher on-state power loss, which is deemed unacceptable. Another commonly used TSEP is the turn-on delay time ($t_{d-on}$), known for its high linearity [7-8]. Its sensitivity can be enhanced through an increase in gate resistance ($R_g$) [9]. However, it compromises the switching speed and increases the switching power loss. A previous study introduced a high-sensitivity transient TSEP called turn-on current overshoot $\Delta i_{ds,on}$ [10]. However, the $T_j$ monitoring method based on $\Delta i_{ds,on}$ is limited to silicon carbide (SiC) metal-oxide-semiconductor field effect transistors (MOSFETs) with body diodes due to the reliance on reverse recovery current. For insulated-gate bipolar transistors (IGBTs), the turn-off voltage overshoot $\Delta V_{ce,off}$ serves as a transient TSEP [11] whose sensitivity depends on stray inductance in the bond wire loop. Therefore, advancements in packaging techniques will result in lower sensitivity of $\Delta V_{ce,off}$. There are other TSEPs such as threshold voltage $V_{TH}$ and leakage current that cannot have their sensitivities adjusted [12-13]. To summarize, improving TSEP sensitivity for higher accuracy monitoring of $T_j$ is generally not recommended from an engineering perspective due to negative effects on converter performance and other limitations.

The broad category of linear regression (LR), which encompasses normal LR [14-15], multiple LR [16-17], and logarithm LR [18], is the most widely utilized regression algorithm for obtaining $F$ in the TSEP method due to its simplicity and high computational efficiency. In order to enhance regression accuracy, some researchers have opted for polynomial fitting methods instead of LR at the expense of computational efficiency [19-20]. With the advancement of multi-feature TSEP methods, machine learning techniques have been introduced. In literature [21], a deep learning algorithm is employed to establish the relationship between $T_j$ and ($V_g$, $R_{ON}$). In literature [22], a convolutional neural network (CNN) is trained to predict $T_j$ based on the turn-off voltage. Additionally, in literature [23], an improved sine fish optimization (ISFO) method optimized support vector machine (SVM) algorithm is proposed to improve the accuracy of $T_j$ estimation. However, if the training data measured in the calibration is erroneous, using existing regression methods will obtain a biased model $F$ or has a significant risk of overfitting which limits the generalization.

In fact, the enhancement of $T_j$ monitoring accuracy can be achieved by implementing modifications to the calibration method, which has been overlooked for a considerable period of time. Most proposed calibration methods primarily focus on streamlining the process [3] and enhancing the convenience in field applications [24-25], without contributing to the overall accuracy. However, there are at least three aspects in calibration which can improve the performance of model $F$: **1)** considering the error caused by the additional temperature difference ($\Delta T_{j\text{-}p}$) between the chip and the temperature control plate; **2)** considering the error caused by the effect of stray parameters and the coupled operating parameters; **3)** considering the random measuring error, especially for a transient TSEP.

Therefore, this paper proposes a high-accuracy calibration method for transient TSEPs, and verifies its efficacy in the case of $V_{TH}$ for SiC MOSFETs. The framework consists of three modules that address the three kinds of errors accordingly, which are also the main three contributions of this article.

1) the thermal effect of the load current during the dual pulse test is theoretically analyzed, and an effective compensating strategy for $\Delta T_{j\text{-}p}$ is proposed. Thus, $T_j$ of the device can be determined with enhanced accuracy.

2) a new mechanism is discovered based on the analysis of stray parameters that there is a non-linear and complicated relationship between the bus voltage $V_{bus}$ and the measured $V_{TH}$, which is ignored in existing studies. Thus, $V_{bus}$ is considered in the model $F$, and the interaction between $V_{bus}$ and $V_{TH}$ is obtained by adopting a neural network. Then, the accuracy of $V_{TH}$-based $T_j$ monitoring method is improved and becomes more robust to operating conditions.

3) the random measuring error is discovered to follow a logarithmic Gaussian distribution based on repeating sampled data, indicating there is an unknown hidden variable $\varepsilon_r$ affecting the calibration result. Thus, the repeating measurement is necessary to uncover the information of $\varepsilon_r$. Increasing the number of measurements allows us to more closely approximate the true distribution of $\varepsilon_r$.

The proposed method is validated by experiments in a DC-DC converter for a $T_j$ estimation mission. Compared with TSEP methods with conventional calibration techniques, the overall accuracy is improved by **30%** without additional hardware cost and invasiveness. Moreover, the proposed calibration method has a good generalization in transient TSEP methods for various types of power devices.

The rest of this paper is organized as follows. In Section II, the principle of conventional calibration methods based on a dual test is illustrated. In Section III, the error existing in conventional calibration methods is illustrated and the proposed calibration method for $V_{TH}$ is introduced. In Section IV, the proposed method is validated by experiments based on a DC-DC converter. Finally, the conclusion is drawn in Section VI.

## II. PRINCIPLE OF CONVENTIONAL CALIBRATION METHOD BASED ON A DUAL PULSE TEST

### A. The setup of a dual pulse test platform

A schematic illustration of the dual pulse test platform (DPTP) is depicted in Fig. 1. (a) The DPTP consists of a half-bridge configuration of a DC-DC converter, where the lower SiC MOSFET $M_2$ functions as the device under test (DUT). $M_2$ is securely mounted on a temperature-controlled heating plate to regulate its junction temperature ($T_j$).

The typical waveforms of the load current $i_L$ and the gate voltage are illustrated in Fig. 1. (b). At $t_0$, $M_2$ is turned on and $V_{bus}$ charges the load inductance $L$ through the load $V_{bus}$-$L$-$M_2$ circuit. From $t_0$ to $t_1$, both the drain-source current of $M_2$ ($i_{ds}$) and $i_L$ increase linearly over time. At $t_1$, $M_2$ turns off, and $i_L$ continues flowing through the body diode of $M_1$ ($D_U$). During the interval [$t_1$, $t_2$], $i_L$ remains nearly constant. At $t_2$, $M_2$ is turned on again, causing a commutation of $i_L$ from $D_U$ to $M_2$. Throughout [$t_2$, $t_3$], $i_L$ increases gradually with time. Finally at $t_3$, as $M_2$ turns off once more, $i_L$ circulates through $L$-$D_U$ and decays slowly until it reaches zero over an extended period of time. This entire process represents a single dual pulse test.

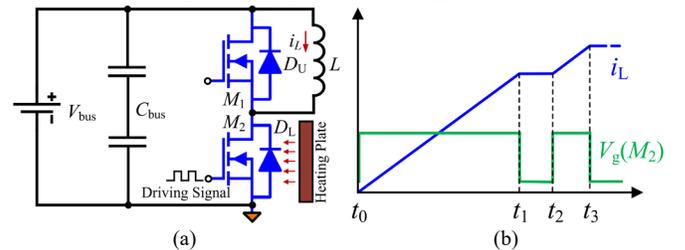

Fig. 1. The principle of dual pulse test, (a) the diagram of dual pulse platform, and (b) the typical waveforms

### B. The principle of a calibration based on the dual pulse test

Fig. 2 displays a conventional calibration process based on the dual pulse test. First, a transient TSEP $\alpha$ is selected. Then, the mechanism of $\alpha$ is analyzed based on semiconductor physics to determine the related parameters, such as the operating conditions $V_{bus}$ and $i_L$. The TSEP $\alpha$ can be expressed as $\alpha = H(T_j, V_{bus}, i_L)$, where $H$ represents a specific multiplicative function. A dataset of [$\alpha$, $V_{bus}$, $i_L$, $T_j$]$_{n\times 4}$ must be established; that is to say that $\alpha$ needs to be discretely measured under each condition of different $V_{bus}$, $i_L$ and $T_j$, and this process is repeated until there is enough data for training the predictive model $F$.

In general, $\alpha$ is measured during the turn-on transient at $t_2$. Since the gap between $t_1$ and $t_2$ is short, $i_L$ at $t_2$ can be can be considered equivalent to $i_L$ at $t_1$. Thus, it can be mathematically represented as:

$$i_L(t_2) = i_L(t_1) = (t_1 - t_0)V_{bus}/L. \quad (1)$$

To maintain a constant $V_{bus}$, $t_1$ is typically adjusted to regulate $i_L(t_2)$ to a specific value. In contrast, regulating $V_{bus}$ and $T_j$ is relatively easier as the former can be manipulated through the DC power source while the latter is considered equivalent to the heating plate's temperature.

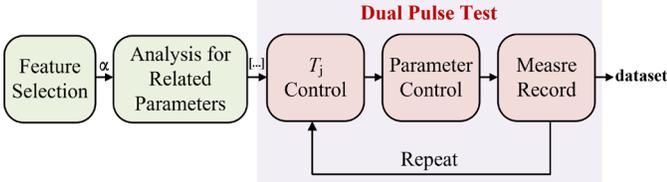

Fig. 2. A conventional calibration framework based on the dual pulse test

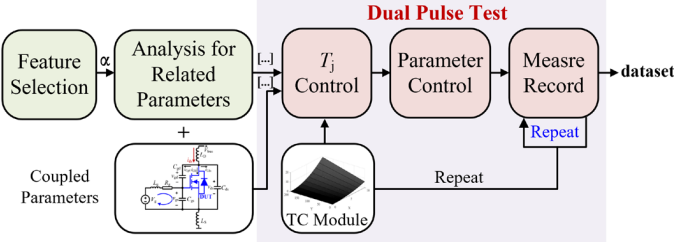

Fig. 3. The proposed calibration framework based on the dual pulse test

### III. THE PROPOSED CALIBRATION METHOD

#### A. The proposed calibration framework

In the introduction, it has been highlighted that conventional calibrations can give rise to significant errors in three aspects as delineated below.

1) Neglecting the additional temperature difference ($\Delta T_{j-p}$) caused by the load current between the chip and the heating plate results in an error when considering its temperature as $T_j$.

2) The impact of stray parameters and coupled parameters is disregarded. Taking $V_{TH}$ as an example, it is solely determined by $T_j$ based on semiconductor physics principles. However, during actual measurements, the effect of $V_{bus}$ becomes coupled through $C_{ds}$. In such cases, the calibration process must consider $V_{bus}$ as a coupled parameter. Consequently, incorporating $V_{bus}$ enhances the robustness of the obtained $F$.

3) There must be some random errors when measuring the TSEP. Especially for transient TSEPs, only a single value can be obtained per measurement. Consequently, the impact of random error is much greater compared with steady-state TSEPs.

Therefore, a new calibration framework is proposed in this paper, as shown in Fig. 3. Compared with conventional calibration framework in Fig. 2, three modules are added. *First*, a temperature compensation (TC) strategy is proposed to eliminate $\Delta T_{j-p}$. *Second*, the coupled parameters are considered according to analyses of stray parameters. *Third*, the transient TSEP is measured repeatedly even under the same condition. A measurable hidden variable is discovered in this process, which can guide the calibration.

#### B. The proposed temperature compensation strategy

Analyses about the thermal effect of $i_L$ forms the basis of the proposed TC strategy. Fig. 4 displays the adjustment of $i_L$ and $V_{bus}$ during the dual pulse test. While one parameter is adjusted, the other remains constant. The load current $i_L$ in the first pulse can be expressed as:

$$i_L(t) = V_{bus}t/L. \quad (2)$$

Thus, the average power loss $P_{loss}$ in the interval of $[0, t_3]$ can be calculated as:

$$P_{loss} = \frac{1}{t_3}\int_0^{t_3} i_L(t)v_{ds,on}(t)dt, \quad (3)$$

where $v_{ds,on}$ is the on-state voltage drop of the DUT (SiC MOSFETs). The ratio of $v_{ds,on}$ and $i_L$ equals to the on-state resistance $R_{ON}$, and thus (3) can be written as:

$$P_{loss} = \frac{R_{ON}}{t_3}\int_0^{t_3} i_L^2(t)dt. \quad (4)$$

Submitting (2) into (4), and the result is

$$P_{loss} = \frac{1}{3}I_3^2 R_{ON}(T_j), \quad (5)$$

where $R_{ON}$ is related to $T_j$.

According to (5), $P_{loss}$ is only determined by the load current and $T_j$, which is unrelated to $V_{bus}$.

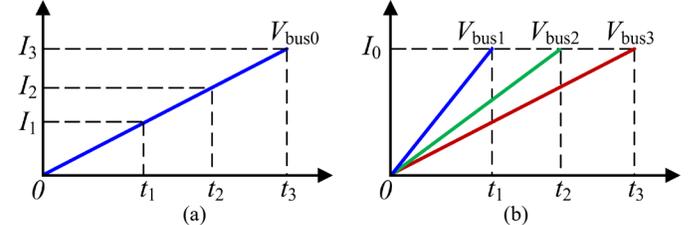

Fig. 4. The current of the first pulse, (a) adjusting $i_L$ and (b) adjusting $V_{bus}$

Based on the thermal circuit theory, $\Delta T_{j-p}$ can be calculated as:

$$\Delta T_{j-p} = P_{loss}R_{th,jp} = \frac{1}{3}I_3^2 R_{ON}(T_j)R_{th,jp}, \quad (6)$$

where $R_{th,jp}$ is the thermal resistance between the chip and the heating plate [27]. Thus, the real temperature of the DUT is

$$T_{re} = T_j + \Delta T_{j-p}. \quad (7)$$

In other words, as long as the temperature of the heating plate and the target load current value are known, the temperature compensation can be achieved according to (7) to improve the calibration accuracy. The efficacy of this approach will be substantiated through a practical example provided subsequently.

The selected DUT is a SiC MOSFET provided by CREE, C2M0080120D. Its $R_{th,jp}$ has been measured in previous work as 0.96 °C/W according to the standard of EIA/JESD51-1 [4]. The relationship between $R_{ON}$ and $T_j$ can be linearized as [4]:

$$R_{ON} = 0.6181T_j + 64.004. \quad (8)$$

The specific measurement process is omitted here. Submitting (8) and the value of $R_{th,jp}$ into (6), a TC function can be obtained as

$$\Delta T_{j\text{-}p} = \frac{1}{3} I_L^2 (0.6181 T_j + 64.004) R_{th,jp}. \quad (9)$$

$I_L$ is the target value of the load current.

The established TC function is displayed in Fig. 5. It can be observed that the self-heating effect of the load current during dual pulse operation has a significant impact, which escalates with both $T_j$ and $I_L$. In essence, Fig. 5 can also represent the inherent error caused in the temperature control loop of conventional calibration methods.

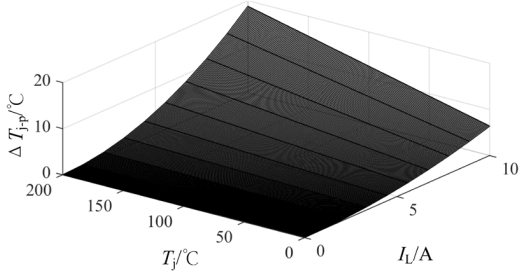

Fig. 5. The established temperature compensation function

### C. The analysis of stray parameters and the consideration of coupled parameters

Taking $V_{TH}$ as an example of transient TSEPs, it can be expressed according to semiconductor physics as

$$V_{TH} = (\phi_{ms} - \frac{Q_f + qN_{it}}{C_{OX}}) + 2\psi_B + \frac{\sqrt{4q\varepsilon_{SiC}N_A\psi_B}}{C_{OX}}, \quad (8)$$

where $\phi_{ms}$ is the working function difference between the metal and the semiconductor, $Q_f$ is the fixed charge in the oxide layer, $C_{OX}$ is the specific capacitance of the oxide layer, $N_{it}$ is the number of trapped charges, $q$ is the basic charge constant, $\Psi_B$ is the Fermi potential, $\varepsilon_{SiC}$ is the relative dielectric constant of SiC, $N_A$ is the doping concentration.

$\Psi_B$ is related to $T_j$, which can be expressed as:

$$\psi_B = \frac{kT_j}{q} \ln \frac{N_A}{n_i}, \quad (9)$$

where $k$ is the Boltzmann constant. $n_i$ is the intrinsic carrier concentration, which can be expressed as:

$$n_i = N_C \exp(-\frac{E_C - E_{Fi}}{kT_j}), \quad (10)$$

where $E_C$ is the conduction energy, $E_{Fi}$ is the intrinsic Fermi level, and $N_C$ is the conduction state density.

Substitute (10) into (9), and $\psi_B$ can be expressed as

$$\psi_B = \frac{kT_j}{q} \ln \frac{N_A}{N_C} + \frac{E_C - E_{Fi}}{q}. \quad (11)$$

For a common N-channel MOSFET, $N_A$ is much lower than $N_C$. Thus, $\psi_B$ decreases with the increase of $T_j$ according to (9).

According to (8)-(11), the threshold voltage is not affected by the operating conditions but solely determined by $T_j$ because $\Psi_B$ increases with $T_j$. However, considering the measurement perspective, there are other parameters affecting $V_{TH}$, which are called couped parameters.

The capturing method in [12] is a commonly used technique for measuring $V_{TH}$. It involves simultaneous measurement of the gate-source voltage $v_{gs}$ and the drain-source current $i_{ds}$. $V_{TH}$ can be determined as the $v_{gs}$ value at time $t_{TH}$ when $i_{ds}$ starts to increase. $t_{TH}$ is usually judged by the voltage induced on the source stray inductance $L_S$, namely $V_{LS}$. When $V_{LS}$ exceeds a small positive threshold $\delta$, it is affirmed that $i_{ds}$ starts to increase. $\delta$ is related to the resolution of the measuring device. This measurement principle can be mathematically expressed as follows:

$$\begin{cases} V_{LS}(t_{TH}) = L_S \frac{di_{ds}}{dt}(t_{TH}) = \delta \\ V_{TH} = v_{gs}(t_{TH}) \end{cases}. \quad (12)$$

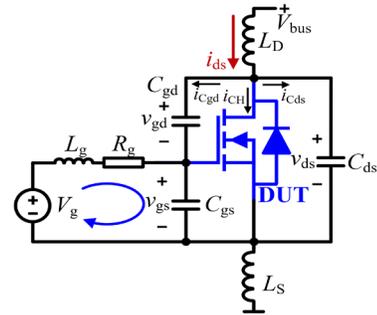

Fig. 6. The principle of the capturing method to measure $V_{TH}$

Then, the effect of stray parameters on the measuring result of $V_{TH}$ can be analyzed.

It can be obtained based on the Kirchhoff's law using in the power loop:

$$\begin{cases} i_{ds} = i_{Cgd} + i_{CH} + i_{Cds} \\ i_{Cgd} = C_{gd} \frac{dv_{gd}}{dt}, i_{Cds} = C_{ds} \frac{dv_{ds}}{dt}, i_{CH} = \frac{v_{ds}}{r_{ds}}, \\ v_{ds} = V_{bus} - (L_D + L_S) \frac{di_{ds}}{dt}, v_{gd} = v_{ds} - v_{gs} \end{cases} \quad (13)$$

where $C_{gd}$ is the equivalent capacitance between gate and drain, $C_{gs}$ is the equivalent capacitance between gate and source, $C_{ds}$ is the equivalent capacitance between drain and source, $i_{Cgd}$ is the current through $C_{gd}$, $i_{CH}$ is the channel current, $i_{Cds}$ is the current through $C_{ds}$, $v_{gd}$, $v_{ds}$, and $v_{gs}$ are the voltages across $C_{gd}$, $C_{ds}$ and $C_{gs}$ respectively, $r_{ds}$ is the transient equivalent resistance between drain and source, $L_D$ and $L_S$ are the stray inductance of drain and source.

The turn-on current changing rate $di_{ds}/dt$ can be expressed as [28]:

$$\frac{di_{ds}}{dt} = \frac{W\mu C_{OX}}{L_{CH}} (1 + \lambda_{CH} V_{bus})(v_{gs} - V_{TH}) \frac{dv_{gs}}{dt}, \quad (14)$$

where $W$ is the cell length, $L_{CH}$ is channel length, $\lambda_{CH}$ is the channel modulation factor, and $\mu$ is the carrier mobility.

Conducting a comprehensive analysis of (12), (13) and (14) can reveal an important conclusion that $V_{bus}$ is coupled into the measuring result of $V_{TH}$ by stray parameters. First, an increase in $V_{bus}$ leads to the increase of $di_{ds}/dt$ according to (14). Meanwhile, $v_{ds}$ and $v_{gd}$ increase. Thus, the main component of $i_{ds}$, namely $i_{CH}$, increases, which further causes a rise in the $i_{ds}$ changing rate. In this case, $V_{LS}$ reaches $\delta$ in a shorter period, where $t_{TH}$ is captured to be smaller and $V_{TH}$ is measured to be lower. Thus, the TSEP feature $V_{TH}$ is related to $V_{bus}$, which is called the coupled parameter.

In summary, there are non-linear complicated interactions between the TSEP features and the coupled parameters. Thus, a neural network is used to approximate $F$. It also shares advantage of high computational speed and relatively high robustness against noise.

The adopted fully-connected neural network structure is illustrated in Fig. 7, comprising an input layer, a hidden layer, and an output layer, using Relu as the activation function. The input layer consists of three defined types of variables: the utilized TSEP feature $\alpha_i$, the directly related parameter $r_i$, and the coupled parameter $p_i$. For instance, if $\alpha_i$ represents $V_{TH}$, then $r_i$ is not present and $p_i$ corresponds to $V_{bus}$. The output layer encompasses one variable, namely, the actual temperature obtained through the proposed TC strategy.

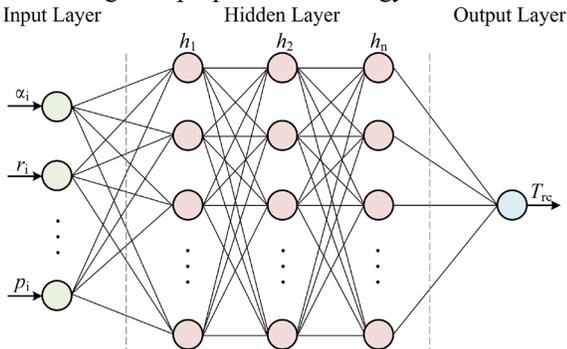

Fig. 7. The adopted neural network structure

For other transient TSEP, it is also necessary to establish a complete stray parameter model combined with semiconductor physics to analyze the coupled parameters. This can improve the robustness of the $T_j$ monitoring method.

### D. The elimination of the random error

When measuring a transient TSEP, the random error is inevitable. Therefore, relying solely on a single measurement in conventional calibration yields considerable errors. In the proposed calibration method, the transient TSEP is measured repeatedly under the same condition to provide a basis for reducing the random error.

Taking $V_{TH}$ as an example, when considering the random error $\varepsilon_r$, it can be established as:

$$V_{TH} = H(T_j, V_{bus}) + \varepsilon_r, \varepsilon_r \sim D, \quad (15)$$

where $\varepsilon_r$ follows a certain distribution $D$. When $D$ is a normal distribution, the effect of $\varepsilon_r$ on the measurement of $V_{TH}$ can be mitigated by repeating the measurement and calculating the mean value. However, based on the distribution fitting result of dual pulse data, it is observed that in most cases $\varepsilon_r$ follows a logarithmic Gaussian distribution. Thus, $\varepsilon_r$ is not a random variable but an unknown and measurable hidden variable. To uncover the information of $\varepsilon_r$, the repeated sampling is necessary.

## IV. EXPERIMENTAL VERIFICATIONS

### A. The experimental setups

The calibration process is conducted based on a dual pulse test platform, as shown in Fig. 8. The DUT used in this process is the SiC MOSFET C2M0080120D. The dual pulse circuit required for testing is provided by Rohm, specifically the P02SCT3040KR-evk-001. The bus voltage necessary for operation is supplied by a DC power source IT6726V. The driving voltage comes from an auxiliary power source HT30-15. A digital signal processor (DSP) embedded in a control board sends driving signals to the system. Probes VAC 4646-X400 and CP9060S are used to measure the load current and drain-source voltage respectively, with their readings displayed on the oscilloscope MSO46. To regulate the junction temperature of DUTs, a heating plate MF7997 is employed, with thermal conductive silicone placed between the DUT and heating plate to minimize thermal resistance.

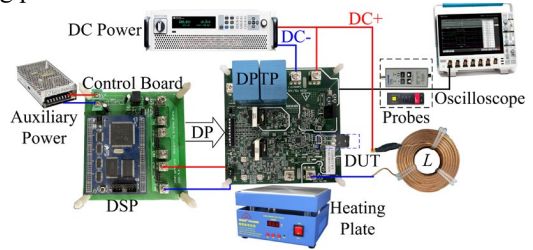

Fig. 8. The dual pulse test platform

The proposed calibration method is validated based on a DC-DC converter, as shown in Fig. 9. Most experimental devices are the same as shown in Fig. 8. An infrared thermal camera provided by Fluke is used to obtain a reference junction temperature.

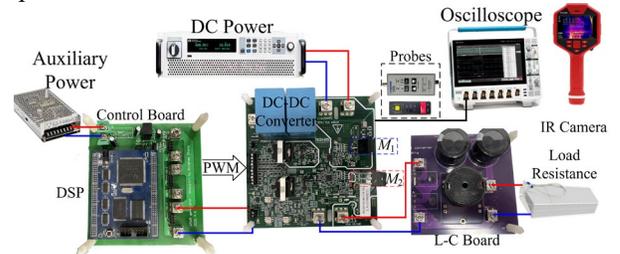

Fig. 9. The DC-DC converter for experimental verification

### B. The experimental verification of the proposed TC strategy

Fig. 10 illustrates the characteristic waveform of a dual pulse test. $V_{TH}$ is mesured based on $V_{LS}$ during the turn-on transient of the second pulse, while the first pulse is used to control $i_L$.

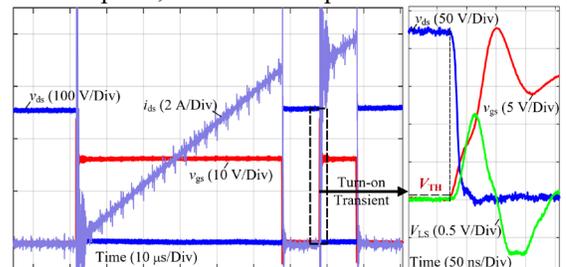

Fig. 10. The typical waveform of dual pulse test and the measurement of $V_{TH}$

Taking $i_L$ = 6 A as an example, the calibration results are displayed in Fig. 11. In this case, model $F$ is derived through linear regression, with the red line indicating the outcome after applying the proposed TC strategy, and the blue line representing the result without the TC strategy. $T_{bc}$ and $T_{ac}$ correspond to the $T_j$ estimated result using these two models. Despite the seemly minor difference between the coefficients, their effects on the $T_j$ monitoring is considerable due to the high sensitivity of the system.

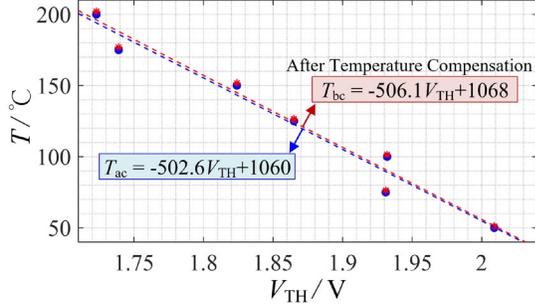

Fig. 11. The calibration result before and after using the TC strategy in the case of $i_L$ = 6A

The proposed TC strategy is verified based on the DC-DC converter displayed in Fig. 9. The bus voltage is set to 300 V, the switching frequency is set to 100 kHz, and the load current is approximately 6 A. The waveforms of related parameters are shown in Fig. 12. (a), where $V_{TH}$ is measured by capturing $V_{LS}$. The reference temperature $T_r$ is measured by the IR camera through a package hole on the DUT. $T_{bc}$ and $T_{ac}$ are obtained from the measured $V_{TH}$ values and the corresponding model $F$. The results are presented in Fig. 12. (b).

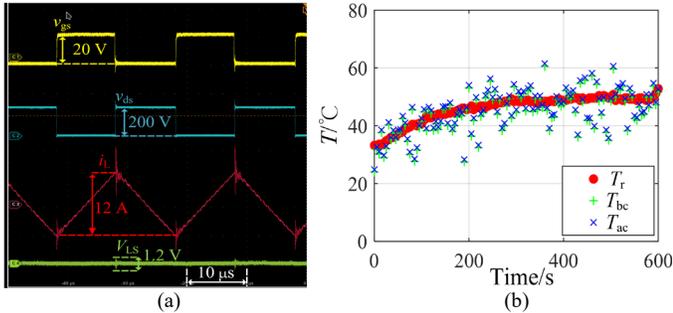

Fig. 12. The experimental results for online $T_j$ monitoring, (a) waveforms of the DC-DC converter, and (b) the monitoring results before and after using the proposed temperature compensation method

The mean absolute error (MAE) of $T_{ac}$ and $T_{bc}$ can be calculated based on Fig. 12. (b), yielding values of 4.29 °C and 4.4 °C, respectively, representing a 2.5% reduction. Similarly, when $i_L$ is 8 A, the MAE reduction is 6.7 %. These results validate the effectiveness of the proposed TC strategy. Furthermore, the performance of the strategy will be higher as the load current increases.

*C. The experimental verification of considering coupled parameters and random errors*

The other two modules of the proposed calibration method are validated together. Different from conventional calibration methods, the TSEP is measured repeatedly $k$ times under the same condition in the proposed calibration framework. To be specific, $k$ is selected as 50 in this research. Fig. 13 displays the repeatedly measuring results of $V_{TH}$ in the case of $i_L$ = 4 A and $V_{bus}$ = 300 V. It can be observed that there is a considerable random error in the measuring results, leading to the fluctuation in the data. Since the data distributions is not symmetric about the mean value, a logarithmic gaussian distribution (LGD) function is used to fit them as shown in Fig. 14. For asymmetric distributions, multiple measurements and averaging to reduce error are no longer effective. Besides, it indicates that there is a hidden variable required to be uncovered in the calibration.

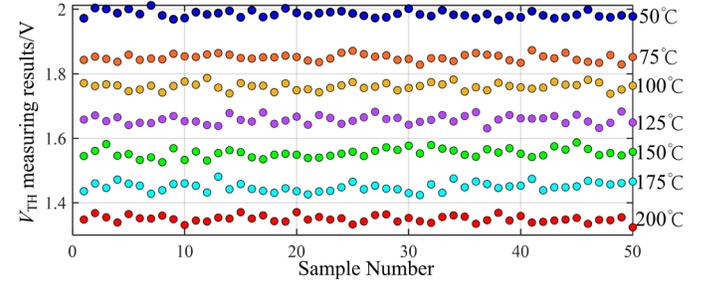

Fig. 13. The repeatedly measuring results of $V_{TH}$ in the case of $i_L$=4 A and $V_{bus}$=300 V

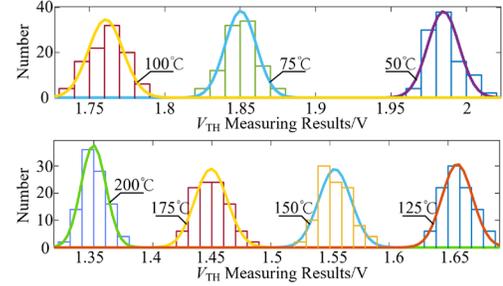

Fig. 14. The distribution fitting results of $V_{TH}$ in the case of $i_L$=4 and $V_{bus}$=300V

Fig. 15 displays the data distribution of three different bus voltage conditions under the same temperature and load current condition ($T_j$ =50 °C, $i_L$ = 4 A). It can be observed that $V_{bus}$ can affect the distribution of the measuring result of $V_{TH}$ ($V_{THm}$). Specially, $V_{THm}$ is negatively related to $V_{bus}$, which is consistent with the theoretical analysis about the effect of stray parameters in Section II. Therefore, when using $V_{TH}$ as a feature to monitor $T_j$, $V_{bus}$ must be added as coupled parameters, which is not considered in conventional calibration methods. Clearly, such neglect can result in significant monitoring errors.

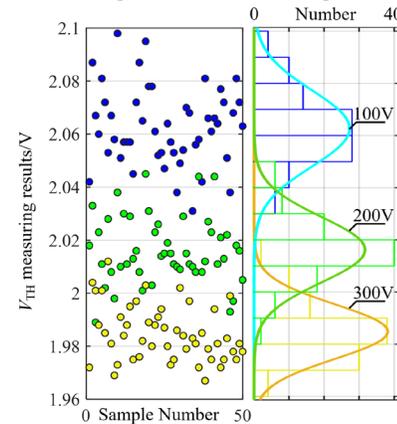

Fig. 15. The data distribution $V_{TH}$ in three different $V_{bus}$ conditions

Fig. 16 displays the measured data of $V_{THm}$ for the proposed calibration method in the case of $i_L$ = 4 A. There are four different bus voltages, and the corresponding compensation temperature is marked. The data after TC will be used to train the model $F$ based on the neural network.

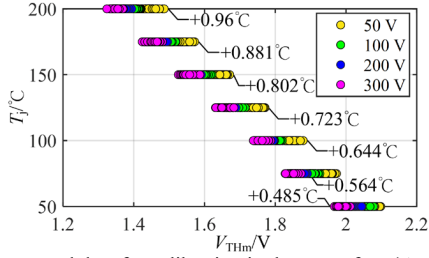

Fig. 16. The measured data for calibration in the case of $i_L$=4A

There are totally 1400 samples in the original dataset, where 15% of them are used as the validation set, 15% of them are used as the test set, and 70% of them are used as the training set. A 10-fold cross verification method is used. In order to balance the accuracy and the computational efficiency, the neural network structure is selected as one hidden layer of 26 neurons. The parameters of the neural network are optimized using the Bayesian regularization method to avoid overfitting to improve the generalization [29]. The corresponding learning curve is displayed in Fig. 17. The mean square error (MSE) converges to below 0.1 in all the three datasets (Training, Validation, Test) after 1000 iterations, as shown in Fig. 17.

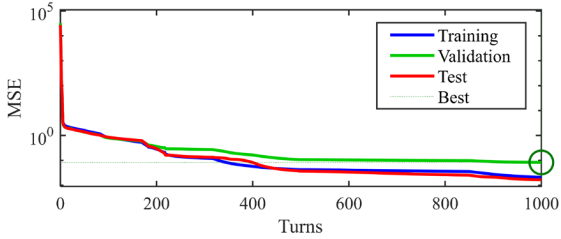

Fig. 17. The measured data for calibration in the case of $i_L$=4A

The error distribution of the trained predictive model $F$ is displayed in Fig. 18. It can be observed that $F$ has a good performance and a good generalization.

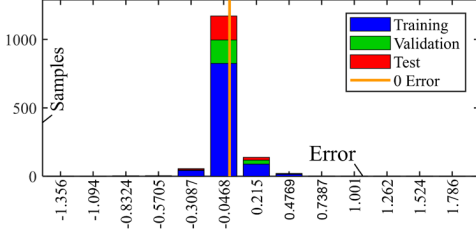

Fig. 18. The measured data for calibration in the case of $i_L$=4A

The obtained model $F$ based on the neural network is visualized and displayed in Fig. 19. Obviously, the monitoring result of $T_j$ ($T_{jm}$) is non-linearly negatively related to $V_{TH}$ and $V_{bus}$, which is consistent with the theoretical analysis in Section II. Thus, in conventional calibration methods, the neglection of coupled parameters such as $V_{bus}$ can result in large monitoring errors.

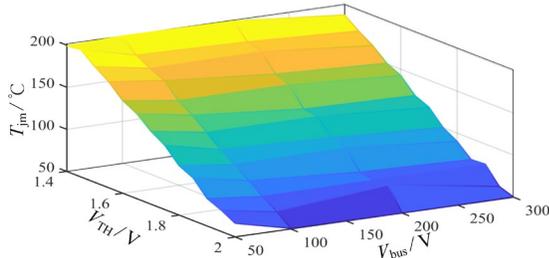

Fig. 19. The visualization of the obtained model $F$ based on the neural network

There are four operating cases of the experiment for junction temperature monitoring as shown in Fig. I. The reference temperature is provided by the IR camera through a hole on the package. The switching frequency is set as 100 kHz. In each case, the DC-DC converter displayed in Fig. 9 operates for 5 minutes from a cold boot, and $T_{jm}$ of the DUT is obtained by the $V_{TH}$ method based on the conventional calibration method ($T_{bc}$-$V_{TH}$ in Fig. 11) and the proposed calibration method respectively. The $V_{TH}$ capturing interval based on $V_{LS}$ is set as 10 seconds.

TABLE I
THE OPERATING CONDITIONS OF THE EXPERIMENT FOR JUNCTION TEMPERATURE MONITORING

| Case | I | II | III | IV |
|---|---|---|---|---|
| $V_{bus}$ | 300 V | 300 V | 200 V | 200 V |
| $i_L$ | 4 A | 6 A | 4 A | 6 A |

Fig. 20 displays the experimental results of $T_j$ monitoring in the four operating cases, where $T_r$ is the reference temperature, $T_{jm1}$ is the monitoring result based on the conventional calibration method, and $T_{jm2}$ is the monitoring result based on the proposed calibration method. It can be observed that the DUT reaches the steady thermal state at 300 s, and the steady junction temperature is related to $V_{bus}$ and $i_L$. In fact, the on-state power loss is determined by $i_L$ and the switching power loss is determined by $V_{bus}$.

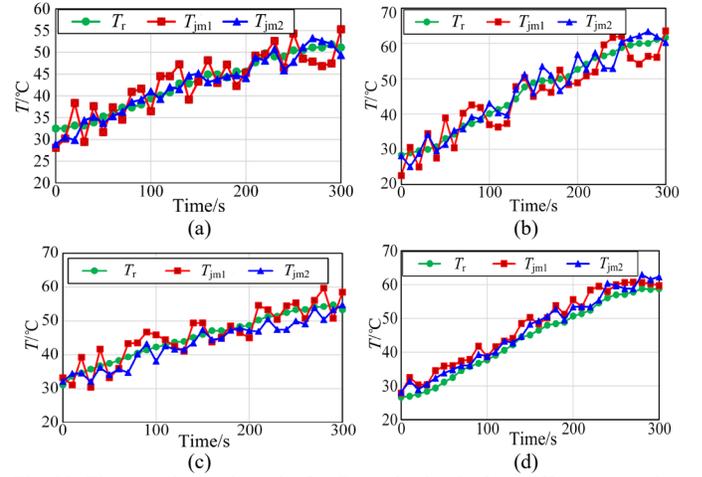

Fig. 20. The experimental results for $T_j$ monitoring in four different operating cases, (a) case I, (b) case II, (c) case III, and (d) case IV

The MAE of $T_j$ monitoring is reduced at least 30% when using the proposed, as shown in TABLE II, where MAE_1 is the mean absolute error of the conventional method and MAE_2 is that of the proposed calibration method. Thus, the effectiveness of the proposed calibration method has been validated. Moreover, the monitoring accuracy does not change with the load current condition. The reason lies in the fact that when measuring $V_{TH}$ based on $V_{LS}$, $i_{ds}$ of the DUT is near zero which is far away from $i_L$. Thus, only considering $V_{bus}$ as coupled parameter in the calibration is reasonable.

TABLE II
REDUCTION OF THE MAE IN THE DIFFERENT OPERATING CASES

| Case | I | II | III | IV |
|---|---|---|---|---|
| MAE_1 | 3.08 | 3.81 | 3.16 | 3.14 |
| MAE_2 | 1.523 | 2.22 | 2.11 | 2.19 |
| Reduction | 50.55% | 41.73% | 33.23% | 30.25% |

It is important to note that while the proposed calibration method has been validated in the case of $V_{TH}$, this does not imply that its applicability is limited to $V_{TH}$. In fact, it can be used in all the transient TSEPs in various types of power devices. The only difference lies in the analyzing of stray parameters and the coupled parameters.

## V. CONCLUSION

A high-accuracy calibration method for transient TSEPs is proposed in this paper. This method addresses three critical issues that limit junction temperature monitoring accuracy in conventional calibration approaches: (1) the additional temperature difference between the device under test (DUT) and the heating plate induced by the load current, (2) the influence of stray parameters and associated coupled parameters such as $V_{bus}$, and (3) the discovered LGD random errors which can not be eliminated by averaging. These issues are addressed by the proposed calibration framework, including a temperature compensation strategy, the stray parameter and coupled parameter analyzing based on an established device model, and the adoption of repeatedly measurement. Finally, the $T_j$ predictive model $F$ is obtained by the neural network. In this manner, the accuracy of junction temperature monitoring can be significantly enhanced, thereby contributing to the reliability of power devices. The proposed calibration method has been validated in a threshold voltage case study based on a DC-DC converter. Experimental results demonstrate that the mean junction temperature monitoring error can be reduced by over 30% in different operating conditions. Compared with alternative methods for improving monitoring accuracy, such as utilizing high-sensitivity TSEPs or modifying fitting regression techniques, the proposed calibration method fundamentally improves the performance of the predictive model $F$ while incurring minimal hardware costs and maintaining strong generalizability.

## REFERENCES

[1] S. Yang, A. Bryant, P. Mawby, D. Xiang, L. Ran and P. Tavner, "An industry-based survey of reliability in power electronic converters", *Proc. IEEE Energy Convers. Congr. Expo.*, pp. 3151-3157, 2009.
[2] Q. Zhang and P. Zhang, "An Online Junction Temperature Monitoring Method for SiC MOSFETs Based on a Novel Gate Conduction Model," *IEEE Trans. Power Electron.*, vol. 36, no. 10, pp. 11087-11096, Oct. 2021.
[3] Y. Yang, X. Ding, G. Sun, G. Lyu and P. Zhang, "IGBT Junction Temperature Monitoring Method Current Calibration Free Based on the Narrow Pulse Injection," *IEEE Trans. Ind. Electron.*, vol. 71, no. 9, pp. 11475-11487, Sept. 2024.
[4] Q. Zhang, G. Lu, Y. Yang and P. Zhang, "A High-Frequency Online Junction Temperature Monitoring Method for SiC MOSFETs Based on on-State Resistance with Aging Compensation," *IEEE Trans. Ind. Electron.*, vol. 70, no. 7, pp. 7393-7405, July 2023.
[5] F. Stella, G. Pellegrino, E. Armando and D. Daprà, "Online junction temperature estimation of SiC power MOSFETs through on-state voltage mapping", *IEEE Trans. Ind. Appl.*, vol. 54, no. 4, pp. 3453-3462, Jul./Aug. 2018.
[6] A. Chanekar, N. Deshmukh, A. Arya and S. Anand, "Gate Voltage-Based Active Thermal Control of Power Semiconductor Devices," *IEEE Trans. Power Electron.*, vol. 38, no. 9, pp. 11531-11542, Sept. 2023
F. Yang, S. Pu, C. Xu and B. Akin, "Turn-on Delay Based Real-Time Junction
[7] Temperature Measurement for SiC MOSFETs With Aging Compensation," *IEEE Trans. Power Electron.*, vol. 36, no. 2, pp. 1280-1294, Feb. 2021.
[8] M. Du, Y. Tang, M. Gao, Z. Ouyang, K. Wei and W. G. Hurley, "Online estimation of the junction temperature based on the gate pre-threshold voltage in high-power IGBT modules", *IEEE Trans. Device Mater. Rel.*, vol. 19, no. 3, pp. 501-508, Sep. 2019.
[9] M. Farhadi, R. Sajadi, B. T. Vankayalapati and B. Akin, "Switching Transient-Based Junction Temperature Estimation of SiC MOSFETs With Aging Compensation," *IEEE Trans. Power Electron.*, vol. 39, no. 10, pp. 12424-12434, Oct. 2024.
[10] Q. Zhang, G. Lu and P. Zhang, "A High-Sensitivity Online Junction Temperature Monitoring Method for SiC MOSFETs Based on the Turn-on Drain–Source Current Overshoot," *IEEE Trans. Power Electron.*, vol. 37, no. 12, pp. 15505-15516, Dec. 2022.
[11] Y. Yang and P. Zhang, "In Situ Insulated Gate Bipolar Transistor Junction Temperature Estimation Method via a Bond Wire Degradation Independent Parameter Turn-OFF Vce Overshoot," *IEEE Trans. Ind. Electron.*, vol. 68, no. 10, pp. 10118-10129, Oct. 2021.
[12] X. Jiang et al., "Online Junction Temperature Measurement for SiC MOSFET Based on Dynamic Threshold Voltage Extraction," *IEEE Trans. Power Electron.*, vol. 36, no. 4, pp. 3757-3768, April 2021.
[13] Q. Zhang, G. Lu, T. Meng and P. Zhang, "An online thermal monitoring method for converter system based on the extraction of DC bus leakage current," *CSEE Journal of Power and Energy Systems*.
[14] B. Shi et al., "Junction Temperature Measurement Method for SiC Bipolar Junction Transistor Using Base–Collector Voltage Drop at Low Current," in *IEEE Trans. Power Electron.*, vol. 34, no. 10, pp. 10136-10142, Oct. 2019.
[15] Z. Xu, F. Xu and F. Wang, "Junction Temperature Measurement of IGBTs Using Short-Circuit Current as a Temperature-Sensitive Electrical Parameter for Converter Prototype Evaluation," *IEEE Trans. Ind. Electron.*, vol. 62, no. 6, pp. 3419-3429, June 2015.
[16] M. Farhadi, R. Sajadi, B. T. Vankayalapati and B. Akin, "Switching Transient-Based Junction Temperature Estimation of SiC MOSFETs With Aging Compensation," *IEEE Trans. Power Electron.*, vol. 39, no. 10, pp. 12424-12434, Oct. 2024.
[17] T. Tang, W. Song, K. Yang and J. Chen, "A Junction Temperature Online Monitoring Method for IGBTs Based on Turn-off Delay Time," *IEEE Trans. Ind. Appli.*, vol. 59, no. 5, pp. 6399-6411, Sept.-Oct. 2023.
[18] H. Niu and R. D. Lorenz, "Sensing Power MOSFET Junction Temperature Using Circuit Output Current Ringing Decay," *IEEE Trans. Ind. Appli.*, vol. 51, no. 2, pp. 1763-1773, March-April 2015.
[19] H. Meng, A. Zhu, H. Zuo, H. Luo, Z. Xin and W. Li, "Online Junction Temperature Extraction with Gate Voltage Under Nontrigger Current for High-Voltage Thyristor," *IEEE Trans. Power Electron.*, vol. 38, no. 9, pp. 10574-10578, Sept. 2023.
[20] U. R. Vemulapati, E. Bianda, D. Torresin, M. Arnold and F. Agostini, "A Method to Extract the Accurate Junction Temperature of an IGCT During Conduction Using Gate–Cathode Voltage," *IEEE Trans. Power Electron.*, vol. 31, no. 8, pp. 5900-5905, Aug. 2016.
[21] M. -K. Kim, Y. -D. Yoon and S. W. Yoon, "Actual Maximum Junction Temperature Estimation Process of Multichip SiC MOSFET Power Modules with New Calibration Method and Deep Learning," *IEEE J. Emerg. Sel. Topics Power Electron.*, vol. 11, no. 6, pp. 5602-5612, Dec. 2023.
[22] H. Wang et al., "A Junction Temperature Monitoring Method for IGBT Modules Based on Turn-Off Voltage with Convolutional Neural Networks," *IEEE Trans. Power Electron.*, vol. 38, no. 8, pp. 10313-10328, Aug. 2023.
[23] L. Li, J. Liu, M. -L. Tseng and M. K. Lim, "Accuracy of IGBT Junction Temperature Prediction: An Improved Sailfish Algorithm to Optimize Support Vector Machine," *IEEE Trans. Power Electron.*, vol. 39, no. 6, pp. 6864-6876, June 2024.
[24] W. Lai et al., "In-Situ Calibration Method of Online Junction Temperature Estimation in IGBTs for Electric Vehicle Drives," *IEEE Trans. Power Electron.*, vol. 38, no. 1, pp. 1178-1189, Jan. 2023.
[25] Y. Peng, Q. Wang, H. Wang and H. Wang, "An On-Line Calibration Method for TSEP-Based Junction Temperature Estimation," *IEEE Trans. Ind. Electron.*, vol. 69, no. 12, pp. 13616-13624, Dec. 2022.
[26] X. Fang, P. Sun, C. He and B. Wang, "Online Converter-Level Temperature Estimation for IGBTs Using Proportional Calibration Method," *IEEE Trans. Power Electron.*, vol. 40, no. 1, pp. 157-161, Jan. 2025.
[27] C. Entzminger, W. Qiao, L. Qu and J. L. Hudgins, "Automated Extraction of Low-Order Thermal Model with Controllable Error Bounds for SiC MOSFET Power Modules," *IEEE Trans. Power Electron.*, vol. 39, no. 1, pp. 538-551, Jan. 2024.
[28] Z. Zeng et al., "Influence of gate loop inductance on TSEP-based junction temperature monitoring for IGBT", *IEEE J. Emerg. Sel. Topics Power Electron.*, vol. 9, no. 4, pp. 4072-4081, Aug. 2021.
[29] W. Li, W. Zhang, Q. Zhang, X. Zhang and X. Wang, "Weakly Supervised Causal Discovery Based on Fuzzy Knowledge and Complex Data Complementarity," *IEEE Trans. Fuzzy Syst.*, vol. 32, no. 12, pp. 7002-7014, Dec. 2024.

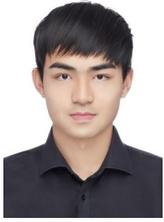

**Qinghao Zhang** (Student Member, IEEE) received the B.Eng. degree and the Ph.D. degree in electrical engineering in 2019 and 2024 from department of electrical engineering, Tsinghua University, Beijing, China, where he is currently working as a "Shuimu" assistant researcher. His current research interests include condition monitoring, health management, diagnostics, and prognostics techniques for power electronic devices in real-time applications. Dr. Zhang has published more than 30 papers, and he was a recipient of three Best Paper Awards from the IEEE IES Society.

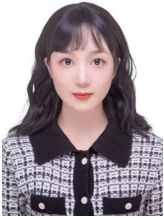

**Wenrui Li** received the B.Eng. degree in automation, in 2020, from the Department of Automation, Sichuan University, Chengdu, China, and the Master's degree in control science and engineering, in 2023, from the Department of Automation, Tsinghua University, Beijing, China, where she is currently working toward the Ph.D. degree in machine learning and pattern recognition. Her current research interests include causal machine learning, applied mathematics and deep learning.

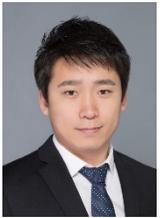

**Pinjia Zhang** (S'06–M'10-SM'17) received the B.Eng. degree in electrical engineering from Tsinghua University, Beijing, China, in 2006 and the Master's and Ph.D. degrees in electrical engineering from Georgia Institute of Technology, Atlanta, GA, USA, in 2009 and 2010, respectively. From 2010 to 2015, he was with the Electrical Machines Laboratory, GE Global Research Center, Niskayuna, NY, USA. Since 2015, he has been with the Department of Electrical Engineering, Tsinghua University as an Associate Professor. His research interests include condition monitoring, diagnostics and prognostics techniques for electrical assets. He has published over 80 papers in refereed journals and international conference proceedings, has over 40 patent fillings in the U.S. and worldwide. Dr. Zhang was the recipient of IAS Andrew W. Smith Outstanding Young Member Achievement Award in 2018. He also received 3 best paper awards from the IEEE IAS and IES society.